%% file: Final.tex
\documentclass[10pt,twocolumn,letterpaper]{article}

\usepackage{cvpr}              

\input{preamble}

\definecolor{cvprblue}{rgb}{0.21,0.49,0.74}
\usepackage[pagebackref,breaklinks,colorlinks,allcolors=cvprblue]{hyperref}

\usepackage{graphicx}
\usepackage{multirow} 

\usepackage{amssymb} 
\usepackage{pifont}  
\usepackage{colortbl}
\usepackage{bm}

\usepackage{caption}
\usepackage{subcaption}
\usepackage[accsupp]{axessibility}

\def\paperID{3926} 
\def\confName{CVPR}
\def\confYear{2026}

\newcommand{\cmark}{\ding{51}}%
\newcommand{\xmark}{\ding{55}}%
\newcommand{\tabref}[1]{Table~\ref{#1}}
\newcommand{\figref}[1]{Fig.~\ref{#1}}
\newcommand{\secref}[1]{Sec.~\ref{#1}}
\def\ourdata{SurgBlood}
\def\ourmodel{BlooDet}
\definecolor{mygray2}{gray}{.92}
\definecolor{ourblue}{rgb}{0.847,0.847,1}
\definecolor{ourpink}{rgb}{1,0.8,0.8}

\graphicspath{{./Imgs/}}
\DeclareGraphicsExtensions{.jpg,.pdf,.png}

\def\eg{\emph{e.g.}}
\def\ie{\emph{i.e.}}

\def\etal{{\em et al.}}

\title{Synergistic Bleeding Region and Point Detection in Laparoscopic Surgical Videos}

\author{
Jialun Pei$^{1}$ \quad
Zhangjun Zhou$^{2}$ \quad
Diandian Guo$^{1}$ \quad
Zhixi Li$^{2,3}$ \quad \\
Jing Qin$^{2}$ \quad 
Bo Du$^{4}$\thanks{Corresponding author. (dubo@whu.edu.cn)} \quad
Pheng-Ann Heng$^{1}$ \\ 
[2mm]
$^1$ The Chinese University of Hong Kong \quad
$^2$ The Hong Kong Polytechnic University \quad \\
$^3$ Southern Medical University \quad
$^4$ Wuhan University \\
}

\begin{document}
\maketitle
\input{sec/0_abstract}    
\input{sec/1_intro}

\input{sec/2_related_work}
\input{sec/3_final_dataset}

\input{sec/4_final_method}
\input{sec/5_final_experiment}
\input{sec/6_conclusion}

\section*{Acknowledgements}

This work was supported in part by the Research Grants Council of the Hong Kong Special Administrative Region, China, under Project T45-401/22-N and in part by the Innovative Research Group Project of Hubei Province under Grants  2024AFA017.

{
    \small
    \bibliographystyle{ieeenat_fullname}
    \bibliography{main}
}


\end{document}

%% file: preamble.tex
\usepackage{xcolor}
\usepackage{bm}



%% file: sec/0_abstract.tex
\begin{abstract}
Intraoperative bleeding in laparoscopic surgery causes rapid obscuration of the operative field to hinder the surgical process and increases the risk of postoperative complications.
Intelligent detection of bleeding areas can quantify the blood loss to assist decision-making, while locating bleeding points helps surgeons quickly identify the source of bleeding and achieve hemostasis in time to improve surgical success rates.
To fill the benchmark gap, we first construct a real-world laparoscopic surgical bleeding detection dataset, named~\textbf{\ourdata}, comprising 5,330 frames from 95 surgical video clips with bleeding region and point annotations. 
Accordingly, we develop a dual-task synergistic online detector called~\textbf{\ourmodel}, enabling simultaneous detection of bleeding regions and points in laparoscopic surgery.
The baseline embraces a dual-branch bidirectional guidance design based on Segment Anything Model 2. The mask branch detects bleeding regions through adaptive edge and point prompt embeddings, while the point branch leverages mask memory to induce bleeding point memory modeling and captures point motion direction via inter-frame optical flow.
By coupled bidirectional guidance, our framework explores spatial-temporal correlations while exploiting memory modeling to infer the current bleeding status.
Extensive experiments indicate that our method outperforms 13 counterparts in bleeding detection.
Code and data are available at \href{https://github.com/PJLallen/SurgBlood}{https://github.com/PJLallen/SurgBlood}.
\end{abstract}

%% file: sec/1_intro.tex
\vspace{-10pt}
\section{Introduction}\label{sec:intro}

Minimally invasive surgery has revolutionized clinical healthcare by reducing patient trauma and accelerating postoperative recovery~\citep{varghese2024artificial,pei2025instrument}.
However, intraoperative bleeding is an emergency that significantly impacts surgical safety and efficiency in endoscopic surgery~\citep{guo2025surgical,gallaher2022acute}. 
Rapid changes in the amount and speed of bleeding can severely obscure the surgical field, delaying the surgeon’s response and reducing the success rate of surgery.
More prolonged bleeding increases the risk of organ damage and postoperative complications~\citep{deziel1993complications}.
Thus, utilizing computer-assisted techniques to detect bleeding regions and localize bleeding points holds significant clinical value.
In one respect, detecting bleeding areas can quantify blood loss, providing timely support for intraoperative decision-making.
In another respect, precise bleeding point localization enables surgeons to control hemorrhage promptly to ensure safety.

\begin{figure}[t!]
	\centering
	\includegraphics[width=0.96\linewidth]{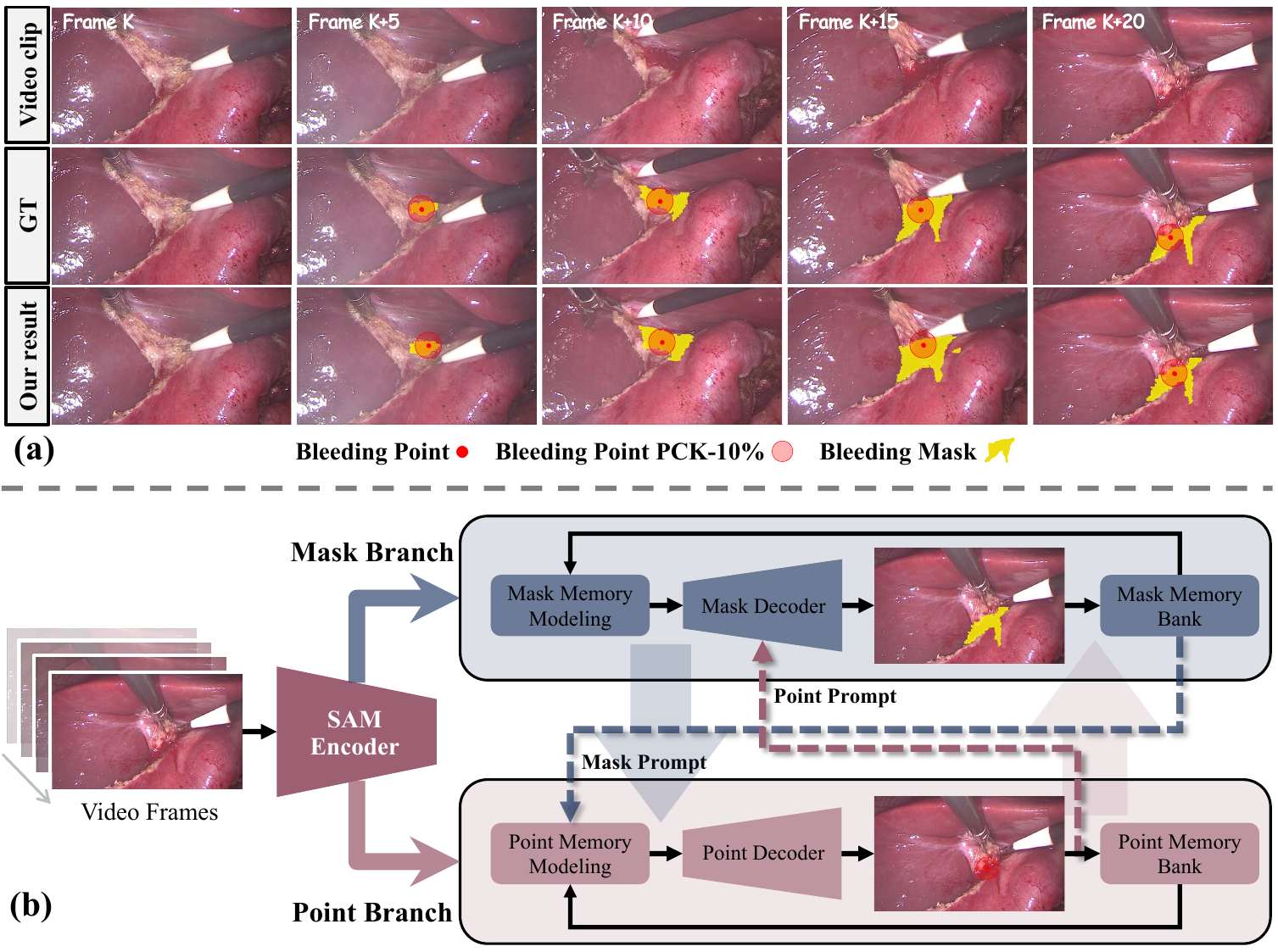}
    \vspace{-0.5em}
	\caption{(a): Illustration of bleeding detection task with samples in~\ourdata~and predictions of our solution. (b): The proposed~\ourmodel~performs dual-branch bidirectional guidance for synergistic bleeding region and point detection.}
\label{Bleed_fig1}
\vspace{-10pt}
\end{figure}

Despite the popularity of laparoscopic surgery, automated detection of bleeding regions and bleeding points still faces numerous challenges~\cite{sunakawa2024deep}.
Due to the narrow field of view under laparoscopy and unstable lighting conditions, the anatomical structures are incompletely exposed, which increases the difficulty of extracting discriminative representations.
Additionally, the rapid accumulation and flow of blood can change tissue appearance and infiltrate surrounding tissues, reducing the availability of low-level visual clues and complicating the detection of bleeding regions. 
The bleeding points may also be buried by blood or obscured by tissues, making it difficult to quickly locate and continuously track~\cite{mori2024red}.
Beyond these challenges, intelligent bleeding warning involves detecting bleeding regions and locating bleeding points during dynamic surgical procedures~\cite{hirai2022evaluation}. 
This requires a reliable multi-task online detector that models fine-grained spatial-temporal relationships in surgical videos for accurate predictions.
Further, the lack of publicly available multi-task real bleeding datasets remains a major obstacle to progress in surgical intelligence.

To advance research on bleeding region and point detection in surgical videos, we construct a new actual laparoscopic surgery bleeding dataset, named~\textbf{\ourdata}.
Our dataset comprises a total of 5,330 video frames from 95 laparoscopic video clips, encompassing multiple types and intensities of bleeding during surgery.
As displayed in~\figref{Bleed_fig1}(a), \ourdata~also provides pixel-level annotations of bleeding regions and bleeding point coordinates by hepatobiliary surgeons, supporting the joint detection of bleeding regions and points.
We evaluate several task-relevant methods on~\ourdata~to establish a comprehensive benchmark for laparoscopic intraoperative bleeding detection, driving further research in surgical assistance.

Existing learning-based methods have been demonstrated to be effective in bleeding region detection~\cite{li2020deep,bourbakis2005neural}.
However, most algorithms~\cite{bourbakis2005neural,li2020deep,das2023multi} are designed for image or keyframe analysis, lacking the ability to model temporal dependencies in surgical videos.
In addition, previous methods~\cite{su2022spatio,mao2024pitsurgrt,sunakawa2024deep} mainly focus on bleeding region detection, which falls short of addressing the clinical needs in locating the bleeding source.
With the emergence of large vision models, Segment Anything Model 2 (SAM 2)~\cite{ravi2024sam} unleashed powerful visual representation capabilities for video sequence modeling. Subsequently, a series of SAM 2-based frameworks~\cite{liu2024surgical,pei2024evaluation,liu2025resurgsam2}  have been proposed in the clinical medicine domain, but have not yet been unified into a multi-task paradigm.
Multi-task frameworks~\cite{das2023multi,mao2024pitsurgrt} can detect both regions and keypoints, but they neglect spatial-temporal modeling, leading to difficulties in stably tracking the movement of bleeding sources.
Moreover, current multi-task architectures are unable to fully harness the potential of SAM 2 in cross-task joint optimization.

To meet the clinical demand of bleeding region and point detection, in this paper, we propose a dual-task online baseline model called~\textbf{\ourmodel}, which embraces a dual-branch bidirectional guidance scheme based on SAM 2 to synergistically optimize both tasks.
As illustrated in~\figref{Bleed_fig1}(b), our framework consists of two branches: \emph{Mask branch} and \emph{Point branch}.
In the mask branch, we embed an edge generator that performs multi-scale perception of spatial-temporal features with the wavelet Laplacian filter to generate edge prompts, mitigating the problem of blurred bleeding boundaries.
Meanwhile, we incorporate bleeding points produced from the point branch as point prompts and combine them with edge prompts to facilitate bleeding region detection.
For the point branch, considering the movement of bleeding points within the field of view is influenced by the relative motion of the camera, we leverage inter-frame optical flow and mask memory to estimate camera motion and viewpoint offsets, improving the location accuracy of bleeding points and reducing interference caused by surrounding blood blurring.
Further, we integrate mask memory trails from the mask branch to enhance bleeding point perception.
By coupling clues and co-guiding across two branches, \ourmodel~can exploit spatial-temporal associations between bleeding regions and points. 
Extensive experiments demonstrate that our approach achieves superior performance in both tasks, \eg, 64.88\% IoU for bleeding region detection and 83.69\% PCK-10\% for bleeding point detection.
The main contributions are four-fold:
\begin{itemize}

\item We debut intraoperative bleeding region and point detection tasks in surgical videos and contribute a real-world dataset to advance the surgical intelligent assistance.

\item We introduce a dual-task synergistic online detector \ourmodel, which adopts a dual-branch structure and performs co-optimization by mutual prompts and bidirectional guidance for bleeding region and point detection.

\item The inter-frame optical flow and mask memory are utilized in point branch for capturing movement cues and providing spatial-temporal modeling. 
Besides, the edge generator and point prompting strategy in the mask branch exploits multi-scale wavelet Laplacian filters to enhance edge perception, while incorporating bleeding points for mask prompt embedding.

\item We establish a comprehensive benchmark and evaluate 13 task-related models on~\ourdata.
Experimental results indicate that our method achieves superior performance for joint bleeding region and point detection.

\end{itemize}

%% file: sec/2_related_work.tex
\section{Related Work}

\noindent\textbf{Bleeding Region Detection.} 
Bleeding region detection has been explored across various medical scenarios, such as intracranial hemorrhage detection~\cite{li2020deep}, capsule endoscopy bleeding recognition~\cite{bourbakis2005neural}, and retinal hemorrhage identification~\cite{wu2024automatic}. 
Deep learning-based methods employ convolution and attention mechanisms to extract discriminative features for bleeding localization. 
Sunakawa \etal~\cite{sunakawa2024deep} developed a semantic segmentation model for automatically recognizing bleeding regions on the anatomical structure of the liver.
Nonetheless, existing methods primarily focus on mask-level bleeding detection, overlooking the localization of the bleeding source.
To bridge this gap, we introduce a unified paradigm for the synergistic detection of bleeding regions and points in surgical videos.

\noindent\textbf{Keypoint Detection in Medical Domain.}
Keypoint detection plays a crucial role in various clinical applications, \eg, pathological site identification and anatomical landmark localization~\cite{vorontsov2024foundation, cui2026depth}. 
Existing keypoint detection methods are usually classified into three categories: 1) Context-aware spatial methods. This technique exploits the stability and uniformity of keypoint spatial distributions to improve localization accuracy~\cite{zhang2020headlocnet,liu2019misshapen, pei2024depth,cui2025topology}. 
2) Multi-stage learning strategies. These architectures follow a coarse-to-fine process to refine keypoint localization through gradual optimization and integrating shallow and deep layers~\cite{dai2024cephalometric,pei2022osformer,zhong2019attention}. 
3) Multi-task learning frameworks, which jointly optimize image segmentation and keypoint detection by constructing a union network~\cite{zhang2020context,das2023multi,mao2024pitsurgrt}.
To enhance mutual guidance between tasks~\citep{pei2024calibnet}, we propose a collaborative dual-branch model that couples bleeding region and point detection.

\begin{figure}[t!]
\centering
\includegraphics[width=\linewidth]{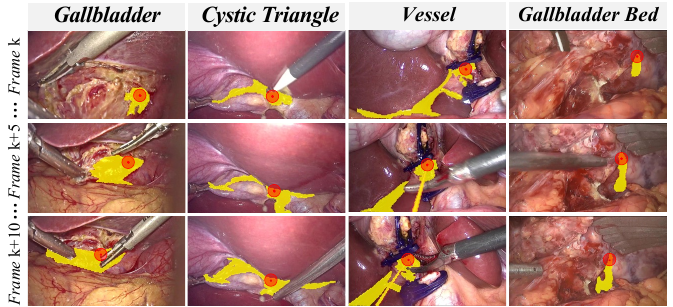}
\caption{Illustration of bleeding types in~\ourdata. 
}
\label{dataExamples}
\vspace{-5pt}
\end{figure}

\begin{table}[t!]
\centering
\scriptsize
\renewcommand{\arraystretch}{1.1}
\setlength\tabcolsep{2.5pt}
\caption{
Summary of publicly available surgical bleeding datasets. SurgBlood provides richer real-world surgical videos with \textit{region} and \textit{point}-level annotations, showing holistic bleeding conditions.
}
\vspace{-0.5em}
\begin{tabular}{lcccccccc}
\toprule
\textbf{Dataset} & \textbf{Year} & \textbf{Source} & \textbf{Type} & \textbf{\#Videos} & \textbf{\#Frames} & \textbf{Region} & \textbf{Point} \\
\midrule
Rabbani \etal~\cite{rabbani2022video} & 2022 & Human & Image & -- & 751 & \checkmark & -- \\
HemoSet~\cite{miao2024hemoset} & 2024 & Animal & Video & 10 & 857 & \checkmark & -- \\
\rowcolor{gray!15}
\textbf{SurgBlood (Ours)} & 2025 & Human & Video & \textbf{95} & \textbf{5,330} & \checkmark & \checkmark \\
\bottomrule
\end{tabular}
\label{comDataset}
\vspace{-5pt}
\end{table}

%% file: sec/3_final_dataset.tex
\section{\ourdata~Dataset}

Currently, there are few publicly available datasets for surgical bleeding detection, as summarized in~\tabref{comDataset}.
HemoSet~\cite{miao2024hemoset} provides bleeding samples based on live animal robotic surgery, including 857 labeled frames for bleeding areas without bleeding points.
Rabbani \etal~\cite{rabbani2022video} released a dataset of gynecologic laparoscopic surgeries, containing 751 frames of active bleeding rather than video clips.
To this end, we construct a brand-new dataset, \ourdata, specifically for bleeding region and point detection in laparoscopic surgery. 
We provide an overview of our dataset from the following three aspects: data collection, data annotation, and data analysis.


\begin{figure}[t!]
\centering
\includegraphics[width=0.7\linewidth]{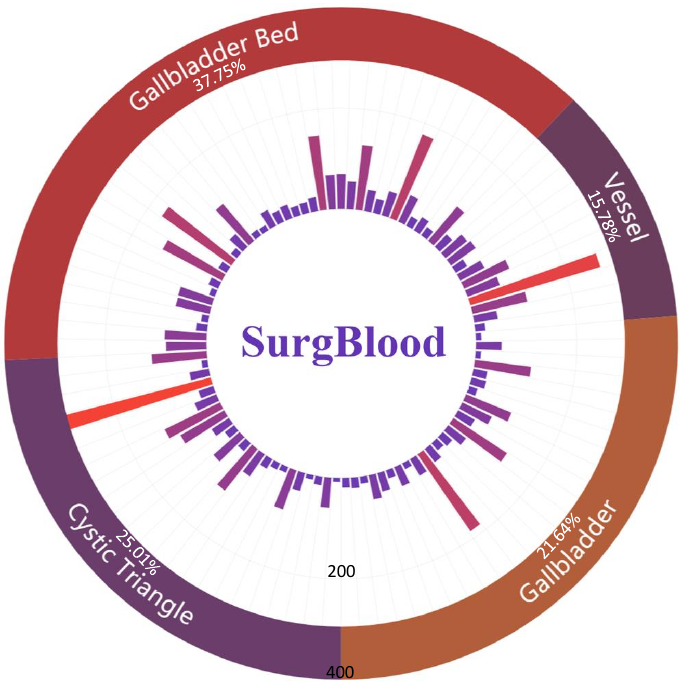}
\caption{Statistical distribution of video clips in~\ourdata.}
\label{data_sta}
\vspace{-5pt}
\end{figure}

\subsection{Data Collection}
To ensure high-quality and representative data, we invited four hepatobiliary surgeons from partner hospitals to carefully select 95 video clips from 42 cholecystectomy surgical cases.
Each clip covers the entire bleeding process while retaining the non-bleeding scene for approximately 3 seconds before and after the bleeding event.
We collect a total of 5,330 video frames with a resolution of 1280$\times$720 from all clips using a sampling rate of 2 fps.
Notably, we focus exclusively on dynamic bleeding regions within the surgical action field, as this is the critical location that directly interferes with the surgeon and contains key bleeding points.
As shown in~\figref{dataExamples}, there are four bleeding types by tissue location: gallbladder, cystic triangle, vessel, and gallbladder bed.
The proportion of bleeding types corresponds to the frequency distribution observed in actual surgery.

\begin{figure}[t!]
\centering
\includegraphics[width=\linewidth]{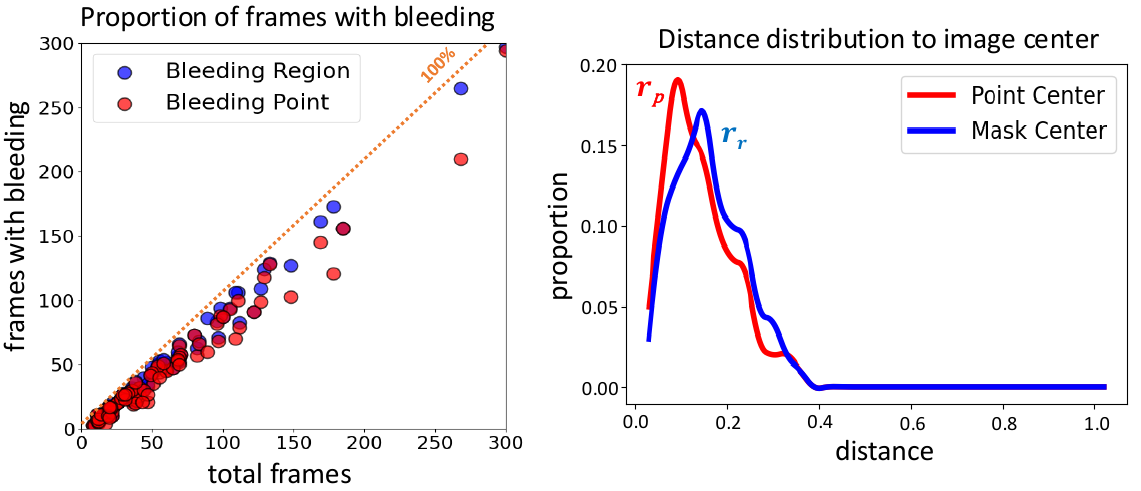}
\caption{\textbf{Bleeding distribution}. Left: proportion of frames with bleeding region and point; Right: Distance of bleeding region center and bleeding point to image center.}
\label{data_his}
\vspace{-5pt}
\end{figure}

\begin{figure*}[t!]
\centering
\includegraphics[width=\linewidth]{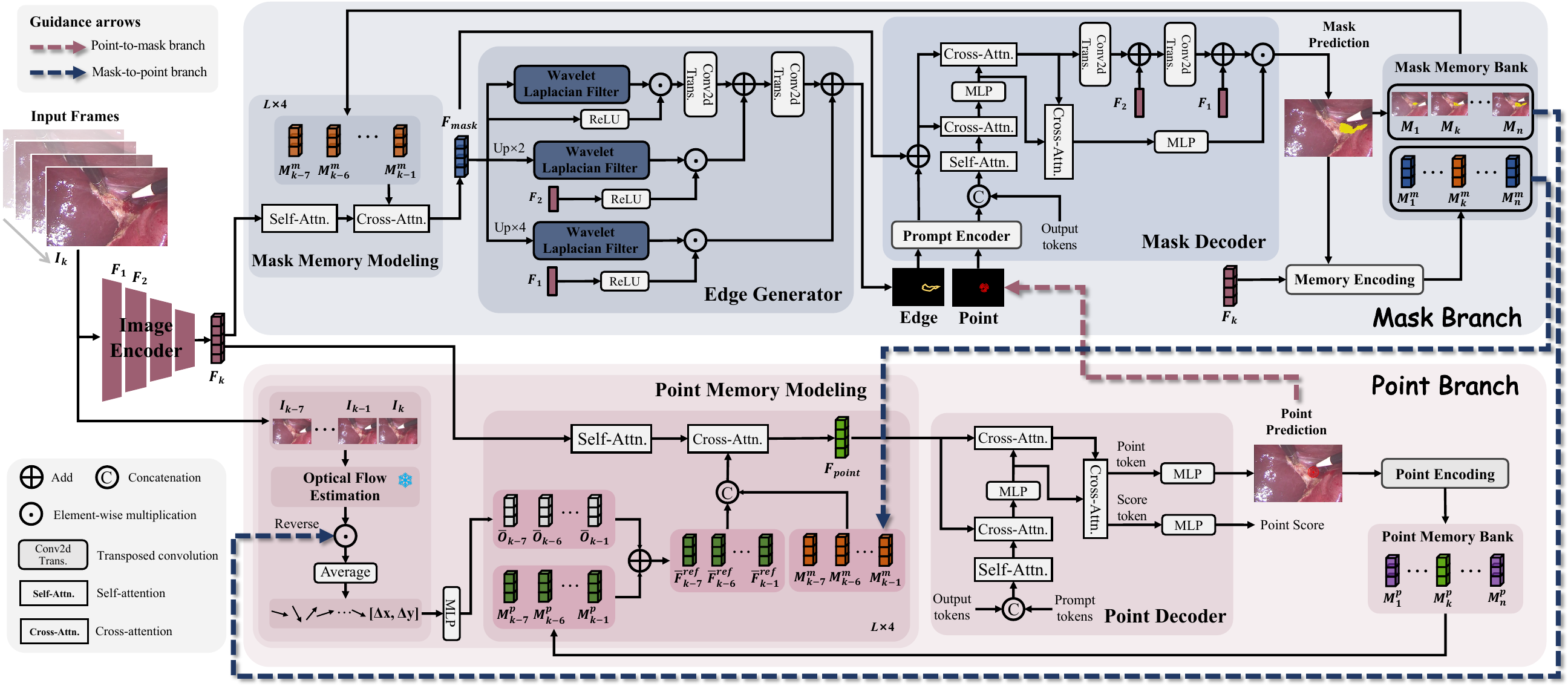}
\caption{Overview of the proposed~\ourmodel. Our framework comprises a mask branch and a point branch to jointly detect bleeding regions and bleeding points. Cross-branch guidance and adaptive prompt embedding allow our framework to reach a co-optimized state.}
\label{framework}
\vspace{-5pt}
\end{figure*}

\subsection{Data Annotation}
To ensure the annotation quality of~\ourdata, the invited hepatobiliary surgeons meticulously annotate and review each video clip.
During the labeling process, the surgeon uses both static frames and dynamic video sequences to label bleeding regions and bleeding point coordinates for each frame.
Annotation is guided by the following principles:
1) For bleeding regions, pooled blood and inactive sparse bloodstains are not annotated.
2) For bleeding points, if the bleeding point is not obscured by tissue or instruments, its coordinates are labeled; if coordinates are difficult to confirm, refer to surrounding anatomy or previous frames to label it.
To ensure annotation consistency, we adopt the cross-validation strategy: each clip is initially annotated by four surgeons, followed by a review and refinement process conducted by two additional surgeons.
\figref{dataExamples}~presents examples of annotations for various bleeding situations.

\subsection{Data Analysis}

\begin{itemize}
\item \emph{\textbf{Clip Distribution}}: 
\ourdata~includes 5,330 frames extracted from 95 video clips. As shown in~\figref{data_sta}, each clip contains an average of 56 frames, with the longest containing 300 frames and the shortest containing 8 frames.
We also counted the distribution of bleeding types: gallbladder (21.64\%), cystic triangle (25.01\%), vessel (15.78\%), and gallbladder bed (37.75\%).

\item \emph{\textbf{Bleeding Ratio}}: 
We calculate the proportion of frames containing bleeding regions and points.
As shown on the left of~\figref{data_his}, both bleeding regions and points have a high frame rate, where the slightly higher rate with bleeding regions is due to the partial occlusion of bleeding points.

\item \emph{\textbf{Space Statistics}}: 
We analyze spatial distribution of bleeding regions and points across all samples in~\ourdata. The right of~\figref{data_his} provides statistical insights into the distances of bleeding region centers and bleeding points from the image center. 
We observed that bleeding primarily occurred in the central zone of surgical manipulation.
\end{itemize}

%% file: sec/4_final_method.tex
\section{Proposed Method}

\ourmodel~is a dual-task collaborative detector for simultaneous bleeding region and point detection in surgical videos. 
The region detection and point localization tasks are tightly coupled: point localization enforces fine-grained geometric consistency within each predicted region (suppressing loose or false detections), while region detection supplies instance-level spatial context that constrains and stabilizes keypoint regression in cluttered scenes. 
This mutual dependency implies that the two tasks constrain and optimize each other rather than being solved in isolation.
We formalize this interdependence with a coupled objective in which parameters of one branch impact the optimization of the other:
\begin{equation}
\{\bm{\theta}^{*}, \bm{\vartheta}^{*}\}
= \arg\min_{\bm{\theta},\,\bm{\vartheta}}
\Big[
    \mathcal{L}_{\mathtt{m}}\big(\bm{\theta}\!\left(\bm{\vartheta}\right)\big)
    + 
    \mathcal{L}_{\mathtt{p}}\big(\bm{\vartheta}\!\left(\bm{\theta}\right)\big)
\Big],
\label{eq:coupled}
\end{equation}
where $\bm{\theta}$ and $\bm{\vartheta}$ denote the model parameters of the region detection and the point localization networks, respectively. The objectives $\mathcal{L}_{\mathtt{m}}$ and $\mathcal{L}_{\mathtt{p}}$ are the training losses associated with these two tasks.

\subsection{Overall Architecture}

As shown in~\figref{framework}, our framework empowers edge clues to detect bleeding regions and incorporates a point branch to localize bleeding points based on SAM 2~\cite{ravi2024sam}. 
The whole model consists of the following processes: image encoder, mask/point memory modeling, Edge generator, mask/point decoder, and mask/point memory bank.

\noindent\textbf{Image Encoding.} 
Given a set of $N$ video frames, including the current frame $I_{k} \in\mathbb{R}^{{H}\times{W}\times{3}}$ and the previous $N$-1 frames $X=\{I_i\}^{k-1}_{i=k-N+1}$, we first flatten frames and feed them into the image encoder inherited from SAM 2 to produce multi-scale spatial features $F\in\mathbb{R}^{{s}\times{c}\times{N}}$, where $s$ denotes the length of the feature sequence and $c$ is the feature dimension. Then, the output sequential frame features $F_{k-N},...,F_{k}$ are fed into the mask and point branches, respectively, for memory modeling.

\noindent\textbf{Point Branch.} 
As shown in the bottom of~\figref{framework}, the point branch comprises three parts: point memory modeling, point decoder, and point memory bank.
The point memory modeling module embeds optical flow estimation to predict the displacement field between consecutive frames $[I_{k-7},...,I_{k}]$, enabling the inference of laparoscopic camera motion and viewpoint offset during surgery, thereby identifying the movement direction of current bleeding points.
Further, we integrate previous mask memory features \{$M^{m}_{q}$\}$_{q=k-7}^{k-1}$ from mask branch with the corresponding point features to enhance location and temporal perception as well as to narrow the search space for bleeding point coordinates.
We describe this process in detail in~\secref{pointbranch}.

Afterward, the memory-enhanced point feature $F_{point}$ passes through the point decoder to predict the bleeding point.
Different from the upsampling fusion in the mask decoder, we employ learnable output tokens and prompt tokens that interact with $F_{point}$ via self-attention and cross-attention~\cite{carion2020end}, followed by MLP layers to predict point coordinates and confidence scores.
The point memory is stored in the point memory bank for temporal modeling.

\noindent\textbf{Mask Branch.} 
The top of~\figref{framework} illustrates the pipeline of the mask branch, which produces bleeding regions by mask memory modeling and coupling edge and point prompts.
The current frame features $F_{k}$ are first fed into mask memory modeling to perform self- and cross-attention interaction~\cite{vaswani2017attention} with mask memory features from previous frames, producing the spatial-temporal feature $F_{mask}$.
After that, we introduce an edge generator that adopts multi-scale wavelet Laplacian filters to $F_{mask}$ for edge refinement. 
Then, we incorporate high-resolution features from the image encoder to obtain edge maps (detailed description in~\secref{maskbranch}).
Unlike the manual intervention prompts~\cite{kirillov2023segment,ravi2024sam}, we form adaptive prompt embeddings by combining the edge map $E_m$ from the edge generator with the point map $P_m$ from point branch.
Then, the prompt encoder is utilized to yield prompt features $E_p$ and $P_p$:
\begin{equation}
E_p,P_p=\mathcal{P}[E_m,P_m],
\end{equation}
\begin{equation}
E_p=\texttt{Conv}\big(\texttt{LN}(\mathbf{G}(\texttt{Conv}(\texttt{LN}(\mathbf{G}(\texttt{Conv}(E_m))))))\big),
\end{equation}
\begin{equation}
P_p=\mathcal{C}[sin(2\pi(\texttt{Po}(P_m))),cos(2\pi(\texttt{Po}(P_m)))]+\texttt{Le},
\end{equation}
where $\mathcal{P}$ represents prompt encoding, $\mathbf{G}$ denotes the GeLU function, $\texttt{Conv}$ refers to 2$\times$2 convolutions, and $\texttt{LN}$ is layer normalization. Also, $\texttt{Po}$ denotes positional encoding, $\texttt{Le}$ stands for learned embeddings, and $\mathcal{C}[,]$ is the concatenation operation.
We input prompt features along with $F_{mask}$ into the mask decoder and attain the bleeding mask by upsampling and integrating with high-resolution features.
Then, we employ memory encoding to achieve mask memory feature $M^{m}_{k}$ and store it in the mask memory bank.
The mask maps are also updated in the memory bank to provide spatial guidance for bleeding point detection.

\noindent\textbf{Cross-branch Guidance.}
Our framework embraces bidirectional collaborative guidance between masks and point branches, enabling simultaneous optimization of bleeding region and bleeding point predictions.
In mask decoder, we exploit the point map produced by the point decoder as an automatic prompt input. This helps guide the decoder to focus on the target bleeding region while mitigating the interference from residual blood in the surrounding area.
In point memory modeling, the predicted mask maps from previous frames provide temporal and directional cues that can improve the accuracy of bleeding point localization.
Besides, mask memory features are merged with point memory features to induce the point decoder to concentrate on the most likely bleeding areas while mitigating the impact of low-contrast background.
The alternating optimization strategy for cross-branch guidance is found in~\secref{sec_loss}.

\subsection{Point Memory Modeling in Point Branch}\label{pointbranch}
To detect bleeding points in consecutive frames effectively, we embed the point memory modeling module in the point branch to develop temporal clues for point features.
As illustrated in~\figref{framework}, point memory modeling is divided into two steps: 1) combining the optical flow between consecutive frames with region maps to compensate for the relative displacement of the bleeding point caused by camera viewpoint offset; 2) interacting the average camera displacement of previous frames with mask memory features from the mask branch to obtain point memory features.


For the viewpoint offset of the camera, we first utilize the frozen PWC-Net~\cite{sun2018pwc} for optical flow estimation.
Given $N$ frames $\{I_i\}^{k}_{i=k-N+1}$, the optical flow $O_{i}(x, y) \in\mathbb{R}^{{H}\times{W}\times{2}}$ between two consecutive frames can be expressed as $ O_{i}(x, y) = \texttt{PWC-Net}(I_{i-1}, I_{i})$.
Considering the instability of the optical flow in the rapidly changing bleeding region, we reverse the mask map $M_{i}$ from mask branch for each frame and combine with $ O_{i}(x, y)$ to obtain the average viewpoint offset $\bar{O_i}(\Delta x, \Delta y)$:
\begin{equation}
\bar{O}_{i}(\Delta x,\Delta y)=\frac{1}{H\times W}\sum_{X=1}^{H}\sum_{Y=1}^{W}(1-M_i)\cdot O_{i}(x,y).
\end{equation} 
Then, the global offset coordinates $\bar{O_i} \in \mathbb{R}^{2}$ of previous frames can be produced via an MLP layer.

After that, we aggregate point memory features $M^p_i$ of previous frames in the point memory bank with $\bar{O_i}$ and concatenate with mask memory features $M^m_i$ to obtain the mask-guided corrected point features $\bar{F}^{ref}_{i}$.
Lastly, we perform self-attention on $F_k$ and cross-attention with $\bar{F}^{ref}_{i}$ to the memory-enhanced point feature $F_{point}$.
Through the optical flow estimation as well as mask guidance, we model the effective memory of point traits with camera offset.

\subsection{Edge Generator in Mask Branch}\label{maskbranch}
To address low contrast and high noise in surgical fields, we embed an edge generator in the mask branch that generates edge map prompts by combining multi-scale wavelet Laplace filters~\cite{lee1996image} with high-resolution features containing low-level texture cues. 
Concretely, we first input the spatial-temporal features $F_{mask}$ into the Gabor wavelet Laplacian filter to enhance edge structures in the spatial domain. The Gabor wavelet operation on position (x, y) is calculated as
\begin{equation}\small
    \mathcal{G}(x,y;\lambda,\theta,\psi,\sigma,\gamma) = exp{(-\frac{x'^{2} + \gamma^{2} y'^{2}}{2\sigma^2})}exp{(i(\frac{2\pi}{\lambda}x'+\psi))} \ ,
\end{equation}
where $x' = xcos\theta + ysin\theta$, $y' = -xsin\theta + ycos\theta$, $\lambda$ denotes the wavelength, $\theta$ is the orientation angle of the Gabor kernel, $\psi$ is the phase offset, $\sigma$ is the standard deviation of Gaussian function, and $\gamma$ stands for the aspect ratio.
Thus, Laplacian filtering based on the Gabor wavelet is defined as
\begin{equation}
\mathbf{L}_\mathbf{g}(x,y) = \Delta{f(x,y)} \cdot \mathcal{G}(x,y),
\end{equation}
\begin{equation}
\Delta{f(x,y)} = \frac{\partial^{2}f}{x^2} + \frac{\partial^{2}f}{y^2},
\end{equation}
where $\Delta{f(x,y)}$ represents the Laplacian operator in 2D space. Then, we perform an activation operation on $F_{mask}$ and interact with the filtered features to suppress low-confidence signals and preserve refined edge features. The whole process of the edge generator can be described as
\begin{equation}
F^{'}_{mask} = (\texttt{ReLU}(F_{mask})) \odot (\mathbf{L}_\mathbf{g}(x,y) \ast {F_{mask}}),
\end{equation}
where $\odot$ and $\ast$ denote the element-wise multiplication and convolution operation. The feature $F^{'}_{mask}$ represents the output from the edge generator. 
As illustrated in~\figref{framework}, we parallel upsample $F_{mask}$ twice and separately pass through the wavelet Laplacian filter, and then interact with high-resolution features ($F_1$ and $F_2$) to refine bleeding edges.
Finally, the generated edge map is fed into the mask decoder as prompts for bleeding region detection.

\begin{table*}[t!]
\centering
\scriptsize
\renewcommand{\arraystretch}{0.85}
\renewcommand{\tabcolsep}{2.26mm}
\caption{Overall comparison with task-related methods on~\ourdata~test set and HemoSet~\cite{miao2024hemoset} dataset. Scores are reported as mean $\pm$ standard deviation. $\dagger$ and $*$ denote region- and point-level models with an additional point/region prediction head for dual-task learning.}
\label{maintable}
\vspace{-5pt}
\begin{tabular}{l l c c c c c c c r}
\toprule
\multirow{3}{*}{\textbf{Methods}} & 
\multirow{3}{*}{\textbf{Volumes}} & 
\multicolumn{4}{c}{\textbf{Bleeding Region Detection}} & 
\multicolumn{3}{c}{\textbf{Bleeding Point Detection}} &
\multirow{3}{*}{\shortstack{\textbf{Params} $\downarrow$}}\\
\cmidrule(lr){3-6} \cmidrule(lr){7-9}
&  & 
\multicolumn{2}{c}{\textbf{\ourdata}} &
\multicolumn{2}{c}{\textbf{HemoSet~\cite{miao2024hemoset}}} &
\multicolumn{3}{c}{\textbf{\ourdata}} &
\\
\cmidrule(lr){3-6} \cmidrule(lr){7-9}
 &  & 
 \textbf{IoU $\uparrow$} & 
 \textbf{Dice $\uparrow$} &
 \textbf{IoU $\uparrow$} & 
 \textbf{Dice $\uparrow$} &
 \textbf{PCK-2\% $\uparrow$} & 
 \textbf{PCK-5\% $\uparrow$} & 
 \textbf{PCK-10\% $\uparrow$} &
 \\

\midrule
\textbf{Swin-UNet$^\dagger$}~\cite{swinunet}      & ECCV'22  & 
$34.83\text{\tiny\textcolor[gray]{0.55}{\,$\pm$\,0.15}}$ & $51.67\text{\tiny\textcolor[gray]{0.55}{\,$\pm$\,0.20}}$ &
$29.86 \text{\tiny\textcolor[gray]{0.55}{\,$\pm$\,0.12}}$ & 
$45.99  \text{\tiny\textcolor[gray]{0.55}{\,$\pm$\,0.17}}$ &
$1.65\text{\tiny\textcolor[gray]{0.55}{\,$\pm$\,0.12}}$ & $11.70\text{\tiny\textcolor[gray]{0.55}{\,$\pm$\,0.18}}$ & $37.23\text{\tiny\textcolor[gray]{0.55}{\,$\pm$\,0.22}}$ & 27.2M \\

\textbf{SAM$^\dagger$}~\cite{kirillov2023segment} & ICCV'23  & 
$37.94\text{\tiny\textcolor[gray]{0.55}{\,$\pm$\,0.17}}$ & $55.01\text{\tiny\textcolor[gray]{0.55}{\,$\pm$\,0.23}}$ &
$38.81 	 \text{\tiny\textcolor[gray]{0.55}{\,$\pm$\,0.18}}$ & 
$55.91  \text{\tiny\textcolor[gray]{0.55}{\,$\pm$\,0.21}}$ &
$4.61\text{\tiny\textcolor[gray]{0.55}{\,$\pm$\,0.14}}$ & $25.86\text{\tiny\textcolor[gray]{0.55}{\,$\pm$\,0.19}}$ & $60.79\text{\tiny\textcolor[gray]{0.55}{\,$\pm$\,0.21}}$ & 93.9M \\

\textbf{MemSAM$^\dagger$}~\cite{deng2024memsam}   & CVPR'24  & 
$52.84\text{\tiny\textcolor[gray]{0.55}{\,$\pm$\,0.16}}$ & $69.14\text{\tiny\textcolor[gray]{0.55}{\,$\pm$\,0.22}}$ &
$ 46.89 	 \text{\tiny\textcolor[gray]{0.55}{\,$\pm$\,0.17}}$ & 
$ 63.84\text{\tiny\textcolor[gray]{0.55}{\,$\pm$\,0.24}}$ &
$5.27\text{\tiny\textcolor[gray]{0.55}{\,$\pm$\,0.13}}$ & $31.80\text{\tiny\textcolor[gray]{0.55}{\,$\pm$\,0.24}}$ & $64.91\text{\tiny\textcolor[gray]{0.55}{\,$\pm$\,0.18}}$ & 133.5M \\

\textbf{STDDNet$^\dagger$}~\cite{chen2025stddnet} & ICCV'25 &
$47.49\text{\tiny\textcolor[gray]{0.55}{\,$\pm$\,0.18}}$ & $64.39\text{\tiny\textcolor[gray]{0.55}{\,$\pm$\,0.19}}$ &
$43.98 	 \text{\tiny\textcolor[gray]{0.55}{\,$\pm$\,0.16}}$ & 
$ 61.10 \text{\tiny\textcolor[gray]{0.55}{\,$\pm$\,0.26}}$ &
$1.48\text{\tiny\textcolor[gray]{0.55}{\,$\pm$\,0.14}}$ & $13.67\text{\tiny\textcolor[gray]{0.55}{\,$\pm$\,0.17}}$ & $45.80\text{\tiny\textcolor[gray]{0.55}{\,$\pm$\,0.20}}$ & 38.2M \\

\textbf{SAM 2$^\dagger$}~\cite{ravi2024sam} & ICLR'25 & 
$50.93\text{\tiny\textcolor[gray]{0.55}{\,$\pm$\,0.12}}$ & $67.49\text{\tiny\textcolor[gray]{0.55}{\,$\pm$\,0.14}}$ &
$ 56.53 	 \text{\tiny\textcolor[gray]{0.55}{\,$\pm$\,0.19}}$ & 
$72.02 \text{\tiny\textcolor[gray]{0.55}{\,$\pm$\,0.24}}$ &
$12.35\text{\tiny\textcolor[gray]{0.55}{\,$\pm$\,0.17}}$ & $41.68\text{\tiny\textcolor[gray]{0.55}{\,$\pm$\,0.21}}$ & $71.99\text{\tiny\textcolor[gray]{0.55}{\,$\pm$\,0.26}}$ & 81.0M \\ 

\cmidrule(lr){1-10}
\textbf{HRNet$^*$}~\cite{sun2019deep} & CVPR'19 & 
$48.90\text{\tiny\textcolor[gray]{0.55}{\,$\pm$\,0.12}}$ & $65.68\text{\tiny\textcolor[gray]{0.55}{\,$\pm$\,0.17}}$ &
$ 48.07 	\text{\tiny\textcolor[gray]{0.55}{\,$\pm$\,0.18}}$ & 
$ 64.93 \text{\tiny\textcolor[gray]{0.55}{\,$\pm$\,0.26}}$ &
$1.98\text{\tiny\textcolor[gray]{0.55}{\,$\pm$\,0.15}}$ & $15.98\text{\tiny\textcolor[gray]{0.55}{\,$\pm$\,0.18}}$ & $47.28\text{\tiny\textcolor[gray]{0.55}{\,$\pm$\,0.23}}$ & 63.6M \\

\textbf{SimCC$^*$}~\cite{li2022simcc} & ECCV'22 & 
$51.02\text{\tiny\textcolor[gray]{0.55}{\,$\pm$\,0.17}}$ & $67.57\text{\tiny\textcolor[gray]{0.55}{\,$\pm$\,0.20}}$ &
$51.62  \text{\tiny\textcolor[gray]{0.55}{\,$\pm$\,0.19}}$ & 
$	68.09  \text{\tiny\textcolor[gray]{0.55}{\,$\pm$\,0.23}}$ &
$3.62\text{\tiny\textcolor[gray]{0.55}{\,$\pm$\,0.13}}$ & $17.30\text{\tiny\textcolor[gray]{0.55}{\,$\pm$\,0.20}}$ & $48.76\text{\tiny\textcolor[gray]{0.55}{\,$\pm$\,0.16}}$ & 66.3M \\

\textbf{GTPT$^*$}~\cite{wang2024gtpt} & ECCV'24 & 
$41.10\text{\tiny\textcolor[gray]{0.55}{\,$\pm$\,0.14}}$ & $58.26\text{\tiny\textcolor[gray]{0.55}{\,$\pm$\,0.18}}$ &
$50.27  \text{\tiny\textcolor[gray]{0.55}{\,$\pm$\,0.20}}$ & 
$	66.91  \text{\tiny\textcolor[gray]{0.55}{\,$\pm$\,0.22}}$ &
$3.29\text{\tiny\textcolor[gray]{0.55}{\,$\pm$\,0.11}}$ & $15.98\text{\tiny\textcolor[gray]{0.55}{\,$\pm$\,0.25}}$ & $38.55\text{\tiny\textcolor[gray]{0.55}{\,$\pm$\,0.18}}$ & 16.7M \\

\textbf{D-CeLR$^*$}~\cite{dai2024cephalometric} & ECCV'24 & 
$51.30\text{\tiny\textcolor[gray]{0.55}{\,$\pm$\,0.19}}$ & $67.82\text{\tiny\textcolor[gray]{0.55}{\,$\pm$\,0.19}}$ &
$50.85  \text{\tiny\textcolor[gray]{0.55}{\,$\pm$\,0.21}}$ & 
$ 	67.42 \text{\tiny\textcolor[gray]{0.55}{\,$\pm$\,0.25}}$ &
$2.97\text{\tiny\textcolor[gray]{0.55}{\,$\pm$\,0.12}}$ & $24.22\text{\tiny\textcolor[gray]{0.55}{\,$\pm$\,0.24}}$ & $63.92\text{\tiny\textcolor[gray]{0.55}{\,$\pm$\,0.17}}$ & 53.4M \\

\textbf{CalibratedSL$^*$}~\cite{feng2025uncertainty} & TMI'25 & 
$36.96\text{\tiny\textcolor[gray]{0.55}{\,$\pm$\,0.16}}$ & $53.97\text{\tiny\textcolor[gray]{0.55}{\,$\pm$\,0.16}}$ &
$44.59 \text{\tiny\textcolor[gray]{0.55}{\,$\pm$\,0.16}}$ & 
$61.68  \text{\tiny\textcolor[gray]{0.55}{\,$\pm$\,0.19}}$ &
$3.46\text{\tiny\textcolor[gray]{0.55}{\,$\pm$\,0.14}}$ & $15.82\text{\tiny\textcolor[gray]{0.55}{\,$\pm$\,0.22}}$ & $39.70\text{\tiny\textcolor[gray]{0.55}{\,$\pm$\,0.13}}$ & 6.6M \\

\cmidrule(lr){1-10}
\textbf{PAINet}~\cite{das2023multi} & MICCAI'23 &
$44.14\text{\tiny\textcolor[gray]{0.55}{\,$\pm$\,0.14}}$ & $61.24\text{\tiny\textcolor[gray]{0.55}{\,$\pm$\,0.15}}$ &
$53.78  \text{\tiny\textcolor[gray]{0.55}{\,$\pm$\,0.17}}$ & 
$	69.87  \text{\tiny\textcolor[gray]{0.55}{\,$\pm$\,0.21}}$ &
$2.47\text{\tiny\textcolor[gray]{0.55}{\,$\pm$\,0.19}}$ & $15.48\text{\tiny\textcolor[gray]{0.55}{\,$\pm$\,0.26}}$ & $48.43\text{\tiny\textcolor[gray]{0.55}{\,$\pm$\,0.20}}$ & 13.6M \\

\textbf{PitSurgRT}~\cite{mao2024pitsurgrt} & IJCARS'24 &
$30.48\text{\tiny\textcolor[gray]{0.55}{\,$\pm$\,0.15}}$ & $46.72\text{\tiny\textcolor[gray]{0.55}{\,$\pm$\,0.17}}$ &
$27.11  \text{\tiny\textcolor[gray]{0.55}{\,$\pm$\,0.18}}$ & 
$	42.65  \text{\tiny\textcolor[gray]{0.55}{\,$\pm$\,0.24}}$ &
$2.47\text{\tiny\textcolor[gray]{0.55}{\,$\pm$\,0.18}}$ & $13.84\text{\tiny\textcolor[gray]{0.55}{\,$\pm$\,0.23}}$ & $41.68\text{\tiny\textcolor[gray]{0.55}{\,$\pm$\,0.18}}$ & 67.3M \\

\textbf{ConsisTNet}~\cite{mao2025consistnet} & IJCARS'25 &
$40.43\text{\tiny\textcolor[gray]{0.55}{\,$\pm$\,0.16}}$ & $57.59\text{\tiny\textcolor[gray]{0.55}{\,$\pm$\,0.18}}$ &
$49.76  \text{\tiny\textcolor[gray]{0.55}{\,$\pm$\,0.22}}$ & 
$66.45  \text{\tiny\textcolor[gray]{0.55}{\,$\pm$\,0.26}}$ &
$7.09\text{\tiny\textcolor[gray]{0.55}{\,$\pm$\,0.15}}$ & $32.83\text{\tiny\textcolor[gray]{0.55}{\,$\pm$\,0.19}}$ & $68.15\text{\tiny\textcolor[gray]{0.55}{\,$\pm$\,0.21}}$ & 143.5M \\

\rowcolor[RGB]{246,232,232}
\textbf{\ourmodel~(Ours)} & - &
$\textbf{64.88}\text{\tiny\textcolor[gray]{0.55}{\,$\pm$\,0.09}}$ & $\textbf{78.70}\text{\tiny\textcolor[gray]{0.55}{\,$\pm$\,0.09}}$ &
$\textbf{59.62}\text{\tiny\textcolor[gray]{0.55}{\,$\pm$\,0.15}}$ & 
$\textbf{74.70}\text{\tiny\textcolor[gray]{0.55}{\,$\pm$\,0.20}}$ &
$\textbf{18.62}\text{\tiny\textcolor[gray]{0.55}{\,$\pm$\,0.13}}$ & $\textbf{55.85}\text{\tiny\textcolor[gray]{0.55}{\,$\pm$\,0.18}}$ & $\textbf{83.69}\text{\tiny\textcolor[gray]{0.55}{\,$\pm$\,0.23}}$ & 91.6M \\
\bottomrule
\end{tabular}
\vspace{-10pt}
\end{table*}

\subsection{Objective Function}\label{sec_loss}
Synergizing two tasks within a unified model to solve coupled objectives is nontrivial, as updating one set of parameters immediately changes the optimal solution for the other.
To this end, we adopt an alternating optimization scheme: fixing one branch during iteration while iteratively updating the other. 
At iteration $t$, we perform two coordinated steps:

\noindent\textbf{1) Update region detection given the current point branch.} We first fix $\bm{\vartheta}^{(t)}$ and update $\bm{\theta}$ by minimizing the region detection loss, which is conditioned on (and regularized by) the current point branch:
\begin{equation}
\bm{\theta}^{(t+1)}
= \bm{\theta}^{(t)}
- \eta_{\theta} \,
\nabla_{\bm{\theta}} \,
\mathcal{L}_{\mathtt{m}}\Big(\bm{\theta}^{(t)} \big(\bm{\vartheta}^{(t)}\big)\Big).
\label{eq:grad_update_theta}
\end{equation}

\noindent\textbf{2) Update point localization given the updated detector.} We then fix $\bm{\theta}^{(t+1)}$ and update $\bm{\vartheta}$ by minimizing the localization loss under the latest detector context:
\begin{equation}
\bm{\vartheta}^{(t+1)}
= \bm{\vartheta}^{(t)}
- \eta_{\vartheta} \,
\nabla_{\bm{\vartheta}} \,
\mathcal{L}_{\mathtt{p}}\Big(\bm{\vartheta}^{(t)} \big(\bm{\theta}^{(t+1)}\big)\Big).
\label{eq:grad_update_vartheta}
\end{equation}

This alternating strategy enforces mutual adaptation: the detector branch $\bm{\theta}$ is optimized under fine-grained geometric constraints from the current localization branch $\bm{\vartheta}^{(t)}$, while the localization branch $\bm{\vartheta}$ is optimized within the updated spatial context provided by $\bm{\theta}^{(t+1)}$. Repeating Eq.~\eqref{eq:grad_update_theta} and Eq.~\eqref{eq:grad_update_vartheta} drives both sets of parameters toward a joint fixed point that is consistent across region- and point-level predictions, thus approximating the coupled optimum in Eq.~\eqref{eq:coupled}.

The loss $\mathcal{L}_{\mathtt{m}}$ of mask branch consists of the region loss $\mathcal{L}_{\mathtt{r}}$ and edge loss $\mathcal{L}_{\mathtt{e}}$:
\begin{equation}
\mathcal{L}_{\mathtt{m}}=\lambda_{\mathtt{r}}\mathcal{L}_{\mathtt{r}}+\lambda_{\mathtt{e}}\mathcal{L}_{\mathtt{e}}.
\end{equation}
$\mathcal{L}_{\mathtt{r}}$ and $\mathcal{L}_{\mathtt{e}}$ are computed as a combination of Focal loss~\cite{lin2017focal} and Dice loss~\cite{milletari2016v}. 
In addition, the point branch loss $\mathcal{L}_{\mathtt{p}}$ consists of the point loss $\mathcal{L}_{\mathcal{P}}$ and the scoring loss $\mathcal{L}_{\mathtt{s}}$:
\begin{equation}
\mathcal{L}_{\mathtt{p}}=\lambda_{\mathcal{P}}\mathcal{L}_{\mathcal{P}}+\lambda_{\mathtt{s}}\mathcal{L}_{\mathtt{s}}.
\end{equation} 
$\mathcal{L}_{\mathcal{P}}$ employs the smooth L1 loss for point-level supervision.
$\mathcal{L}_{\mathtt{s}}$ is the binary cross-entropy loss for point existence.
$\lambda_{\mathtt{r}}$, $\lambda_{\mathtt{e}}$, $\lambda_{\mathtt{s}}$, and $\lambda_{\mathcal{P}}$ are set to 1, 1, 1, and 0.5.

%% file: sec/5_final_experiment.tex
\section{Experiments}

\subsection{Datasets and Evaluation Metrics}

\textbf{Datasets.} 
Since the tasks of the laparoscopic bleeding region and bleeding point cooperative detection are proposed for the first time, we adopt the~\ourdata~dataset to train and test our method and related comparative methods.
We randomly split a total of 95 video clips into two sets: 75 for training and the remaining 20 for testing.
For further evaluation, we assess the region-level detection performance on HemoSet~\cite{miao2024hemoset} and provide comparative results on the Rabbani \etal~\cite{rabbani2022video} dataset in the supplementary materials.

\noindent\textbf{Evaluation Metrics.}
Following previous studies~\cite{ravi2024sam, das2023multi}, we adopt the Intersection over Union (IoU) and Dice Coefficient (Dice) metrics to evaluate bleeding region detection performance. For bleeding points, the Percentage of Correct Keypoints (PCK) metric is used to measure localization accuracy. 
Unlike the 10\% to 40\% threshold range applied in~\cite{das2023multi,mao2024pitsurgrt} for anatomical structure centroids, we adopt a narrower threshold range of 2\%–10\% to ensure greater assessment, \ie, PCK-2\%, PCK-5\%, and PCK-10\%. This is due to the requirement for higher precision and lower tolerance of bleeding point detection in laparoscopic surgery.

\begin{table}[t!]
    \centering
    \scriptsize
    \renewcommand{\arraystretch}{0.8}
    \renewcommand{\tabcolsep}{3.9mm}
    \caption{Comparison with region-specific methods on~\ourdata.}
    \label{maintableRegion}
    \vspace{-5pt}
    \begin{tabular}{l l c c}
    \toprule
    \textbf{Methods} & \textbf{Volumes} & \textbf{IoU $\uparrow$} & \textbf{Dice $\uparrow$} \\
    \midrule
    \textbf{Swin-UNet}~\cite{swinunet}         & ECCV'22 & 
    $41.31\text{\tiny\textcolor[gray]{0.55}{\,$\pm$\,0.18}}$ & 
    $58.47\text{\tiny\textcolor[gray]{0.55}{\,$\pm$\,0.17}}$ \\

    \textbf{SAM}~\cite{kirillov2023segment}    & ICCV'23 & 
    $40.43\text{\tiny\textcolor[gray]{0.55}{\,$\pm$\,0.13}}$ & 
    $57.49\text{\tiny\textcolor[gray]{0.55}{\,$\pm$\,0.15}}$ \\

    \textbf{MemSAM}~\cite{deng2024memsam}      & CVPR'24 & 
    $55.34\text{\tiny\textcolor[gray]{0.55}{\,$\pm$\,0.23}}$ & 
    $71.28\text{\tiny\textcolor[gray]{0.55}{\,$\pm$\,0.13}}$ \\

    \textbf{STDDNet}~\cite{chen2025stddnet}    & ICCV'25 & 
    $50.42\text{\tiny\textcolor[gray]{0.55}{\,$\pm$\,0.17}}$ & 
    $67.04\text{\tiny\textcolor[gray]{0.55}{\,$\pm$\,0.15}}$ \\

    \textbf{SAM 2}~\cite{ravi2024sam}          & ICLR'25 & 
    $63.51\text{\tiny\textcolor[gray]{0.55}{\,$\pm$\,0.11}}$ & 
    $77.68\text{\tiny\textcolor[gray]{0.55}{\,$\pm$\,0.19}}$ \\

    \rowcolor[RGB]{246,232,232}
    \textbf{\ourmodel~(Ours)} & 
    - & 
    $\textbf{64.88}\text{\tiny\textcolor[gray]{0.55}{\,$\pm$\,0.09}}$ & 
    $\textbf{78.70}\text{\tiny\textcolor[gray]{0.55}{\,$\pm$\,0.09}}$ \\
    \bottomrule
    \end{tabular}
    \vspace{-5pt}
\end{table}

\subsection{Implementation Details}
Our framework is implemented on a single RTX 4090 GPU.
During training, we input eight consecutive frames in an online manner with resolution resized to 512$\times$512 pixels. 
The image encoder of~\ourmodel~is initialized with pre-trained weights from SAM 2\_base~\cite{ravi2024sam}.
Additionally, we utilize a frozen PWC-Net~\cite{sun2018pwc} to compute inter-frame optical flow. No data augmentation is applied during data loading.
The maximum learning rate for the image encoder is set to 5e-6, while other parts are trained with 5e-4. 
We employ the Adam optimizer with a warm-up strategy and linear decay, training for 20 epochs.
During inference, we perform frame-by-frame inference in line with SAM 2.

\subsection{Performance Comparison}
We evaluate 13 task-related methods for bleeding region and point detection to build a comprehensive benchmark on~\ourdata, including multi-task detection models~\cite{das2023multi, mao2024pitsurgrt,mao2025consistnet}, region-level object segmentation methods~\cite{swinunet,kirillov2023segment,ravi2024sam,deng2024memsam,chen2025stddnet}, and point-level detection methods~\cite{sun2019deep,li2022simcc,wang2024gtpt,dai2024cephalometric,feng2025uncertainty}.
Besides, we conduct zero-shot testing on the HemoSet~\cite{miao2024hemoset} dataset using pre-trained models from~\ourdata.

\begin{figure*}[t!]
\centering
\includegraphics[width=0.94\linewidth]{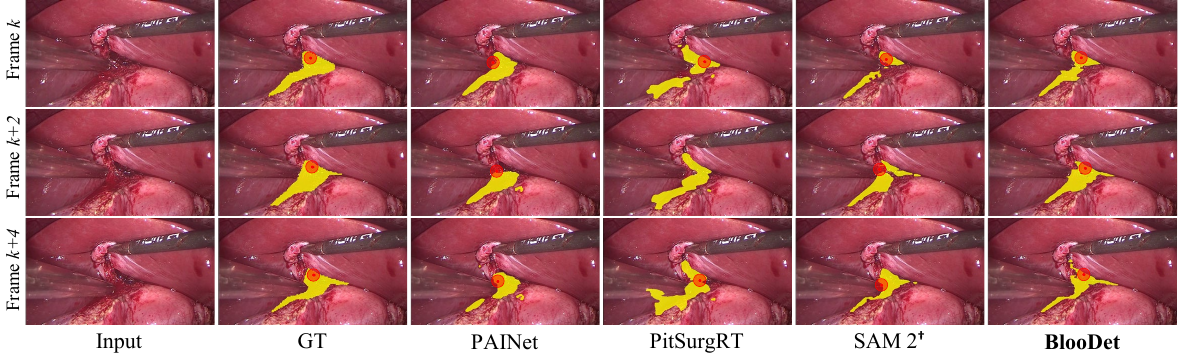}
\vspace{-5pt}
\caption{Visual comparison of bleeding region and point detection on~\ourdata~test set.}
\label{visual_comp}
\vspace{-10pt}
\end{figure*}

\noindent\textbf{Quantitative Evaluation.}
We display the performance of our framework and comparison methods for bleeding region and point detection in \tabref{maintable}.
Region-level and point-level models respectively append point/region prediction heads for dual-task learning.
We can see that our framework outperforms competitors across both tasks on~\ourdata~and 
HemoSet~\cite{miao2024hemoset} datasets. Benefiting from edge-enhanced prompts and bleeding point guidance, \ourmodel~achieves notable improvements in bleeding region detection.
For bleeding point localization, \ourmodel~significantly surpasses other methods thanks to mutual coupling guidance.
Moreover, \ourmodel~obtains superior accuracy with proper parameters.

We also compare with task-specific models on individual tasks. As shown in~\tabref{maintableRegion} and~\tabref{maintablePoint}, \ourmodel~maintains state-of-the-art performance in both bleeding region detection and point localization. 
Comparing the results in~\tabref{maintable} further reveals that task-specific models impact the performance of the original task during dual-task learning, illustrating the effectiveness of our cross-branch guidance and alternating optimization strategy in multi-task learning.

\begin{table}[t!]
    \centering
    \scriptsize
    \renewcommand{\arraystretch}{0.8}
    \renewcommand{\tabcolsep}{1.5mm}
    \caption{Comparison with point-specific methods on~\ourdata.}
    \label{maintablePoint}
    \vspace{-5pt}
    \begin{tabular}{l l c c c}
    \toprule
    \textbf{Methods} & \textbf{Volumes} & \textbf{PCK-2\% $\uparrow$} & \textbf{PCK-5\% $\uparrow$} & \textbf{PCK-10\% $\uparrow$} \\
    \midrule
    \textbf{HRNet}~\cite{sun2019deep}          & CVPR'19 & 
    $3.13\text{\tiny\textcolor[gray]{0.55}{\,$\pm$\,0.17}}$ & 
    $15.98\text{\tiny\textcolor[gray]{0.55}{\,$\pm$\,0.24}}$ & 
    $44.31\text{\tiny\textcolor[gray]{0.55}{\,$\pm$\,0.19}}$ \\

    \textbf{SimCC}~\cite{li2022simcc}          & ECCV'22 & 
    $2.14\text{\tiny\textcolor[gray]{0.55}{\,$\pm$\,0.18}}$ & 
    $14.99\text{\tiny\textcolor[gray]{0.55}{\,$\pm$\,0.22}}$ & 
    $46.95\text{\tiny\textcolor[gray]{0.55}{\,$\pm$\,0.11}}$ \\

    \textbf{GTPT}~\cite{wang2024gtpt}          & ECCV'24 & 
    $2.80\text{\tiny\textcolor[gray]{0.55}{\,$\pm$\,0.15}}$ & 
    $13.01\text{\tiny\textcolor[gray]{0.55}{\,$\pm$\,0.20}}$ & 
    $38.38\text{\tiny\textcolor[gray]{0.55}{\,$\pm$\,0.26}}$ \\

    \textbf{D-CeLR}~\cite{dai2024cephalometric} & ECCV'24 & 
    $5.10\text{\tiny\textcolor[gray]{0.55}{\,$\pm$\,0.16}}$ & 
    $27.67\text{\tiny\textcolor[gray]{0.55}{\,$\pm$\,0.28}}$ & 
    $60.13\text{\tiny\textcolor[gray]{0.55}{\,$\pm$\,0.27}}$ \\

    \textbf{CalibratedSL}~\cite{feng2025uncertainty} & TMI'25 & 
    $3.13\text{\tiny\textcolor[gray]{0.55}{\,$\pm$\,0.15}}$ & 
    $15.82\text{\tiny\textcolor[gray]{0.55}{\,$\pm$\,0.23}}$ & 
    $35.42\text{\tiny\textcolor[gray]{0.55}{\,$\pm$\,0.24}}$ \\

    \rowcolor[RGB]{246,232,232}
    \textbf{\ourmodel~(Ours)} & - & 
    $\textbf{18.62}\text{\tiny\textcolor[gray]{0.55}{\,$\pm$\,0.13}}$ & 
    $\textbf{55.85}\text{\tiny\textcolor[gray]{0.55}{\,$\pm$\,0.18}}$ & 
    $\textbf{83.69}\text{\tiny\textcolor[gray]{0.55}{\,$\pm$\,0.23}}$ \\
    \bottomrule
    \end{tabular}
    \vspace{-10pt}
\end{table}

\noindent\textbf{Qualitative Evaluation.}
\figref{visual_comp} shows a visual comparison of our approach with multi-task methods.
\ourmodel~provides greater stability and consistency in detecting bleeding regions and points.
In surgical environments with low contrast, competitors tend to be disturbed by surrounding noise.
In contrast, our method ensures robust detection across consecutive frames.
More visual comparisons can refer to supplementary materials.


\subsection{Ablation Analysis}

\noindent\textbf{Contributions of Key Component.}
\tabref{table_component}~illustrates the contribution of key components in~\ourmodel.
The experimental results show that point memory modeling contributes significantly to detecting bleeding points, \eg, improving PCK-5\% by 13.02\%. 
Moreover, the edge generator offers effective edge prompt embedding, enhancing the accuracy of bleeding region detection. 
In short, each component makes a positive contribution to model performance.

\noindent\textbf{Ablations for Edge Generator.}
We investigate the effect of edge generator designs in the mask branch.
As exhibited in~\tabref{table_eg}, the wavelet Laplacian filter shows a strong response to edge clues, mitigating interference from complex backgrounds.
Meanwhile, integrating high-resolution features further enhances the quality of edge maps.


\begin{table}[t!]
\centering
\scriptsize
\renewcommand{\arraystretch}{.95}
\setlength\tabcolsep{5.4pt}
\caption{Ablations for key components of~\ourmodel~for bleeding region and point detection on~\ourdata~test set. EG and PMM denote edge generator and point memory modeling modules.}
\vspace{-5pt}
\label{table_component}
\begin{tabular}{cc||cc|ccc}
\hline
 EG & PMM  & IoU $\uparrow$ & Dice $\uparrow$ & PCK-2\% $\uparrow$ & PCK-5\% $\uparrow$ & PCK-10\% $\uparrow$ \\ \hline
\xmark  & \xmark & 56.22 & 71.98 & 13.18 & 45.14 & 75.78 \\\   
 \cmark & \xmark  & 64.38  & 78.36  & 12.52  & 42.83  & 75.94 \\ 
 \xmark & \cmark & 61.20 & 75.93 & 14.33 & 51.57 & 80.89 \\ 
\rowcolor[RGB]{246,232,232}
 \cmark & \cmark  &  \textbf{64.88}  & \textbf{78.70} & \textbf{18.62} & \textbf{55.85} & \textbf{83.69} \\
 \hline
\end{tabular}
\vspace{-5pt}
\end{table}

\begin{table}[t!]
\centering
\scriptsize
\renewcommand{\arraystretch}{0.95}
\setlength\tabcolsep{13pt}
\caption{Influence of edge generator for bleeding region detection.}
\vspace{-5pt}
\label{table_eg}
\begin{tabular}{r|cc|c}
\hline
 Configs  & IoU $\uparrow$ & Dice $\uparrow$ & \# Params (M) \\ \hline
w/o Edge generator  & 61.20  & 75.93 &  \textbf{90.99}  \\
w/o Laplacian Filter & 62.41  & 76.85   &  91.58 \\
w/o $F_{1}\&F_{2}$ & 64.74  & 78.60 &   91.55  \\
\rowcolor[RGB]{246,232,232} 
Edge generator &   \textbf{64.88}  & \textbf{78.70}  & 91.58  \\
\hline
\end{tabular}
\vspace{-5pt}
\end{table}

\begin{table}[t!]
\centering
\scriptsize
\renewcommand{\arraystretch}{0.95}
\setlength\tabcolsep{12.6pt}
\caption{Ablations for optical flow operation via mask maps.}
\vspace{-5pt}
\label{table_optical}
\begin{tabular}{r|ccc}
\hline
 Mask Map  & PCK-2\% $\uparrow$ & PCK-5\% $\uparrow$ & PCK-10\% $\uparrow$  \\ \hline
 w/o optical flow & 14.17	&  45.47 &77.10     \\
 Global & 15.49  &  49.59  &  82.00   \\
Foreground & 12.03  & 41.02   & 71.99   \\
\rowcolor[RGB]{246,232,232} 
Background &  \textbf{18.62} & \textbf{55.85} & \textbf{83.69}  \\
\hline
\end{tabular}
\vspace{-10pt}
\end{table}


\noindent\textbf{Optical Flow Operation Design.}
In point memory modeling, we adopt reversed mask maps in conjunction with optical flow maps to estimate the average camera displacement.
Thus, we ablate the impact of focusing on different regions in mask maps for bleeding point localization. 
\tabref{table_optical}~indicates that using the foreground region leads to inferior performance.
It may stem from the instability of optical flow in the rapidly changing bleeding areas.
In contrast, utilizing the background enables stable motion modeling.

\begin{figure}[t!]
\centering
\includegraphics[width=0.98\linewidth]{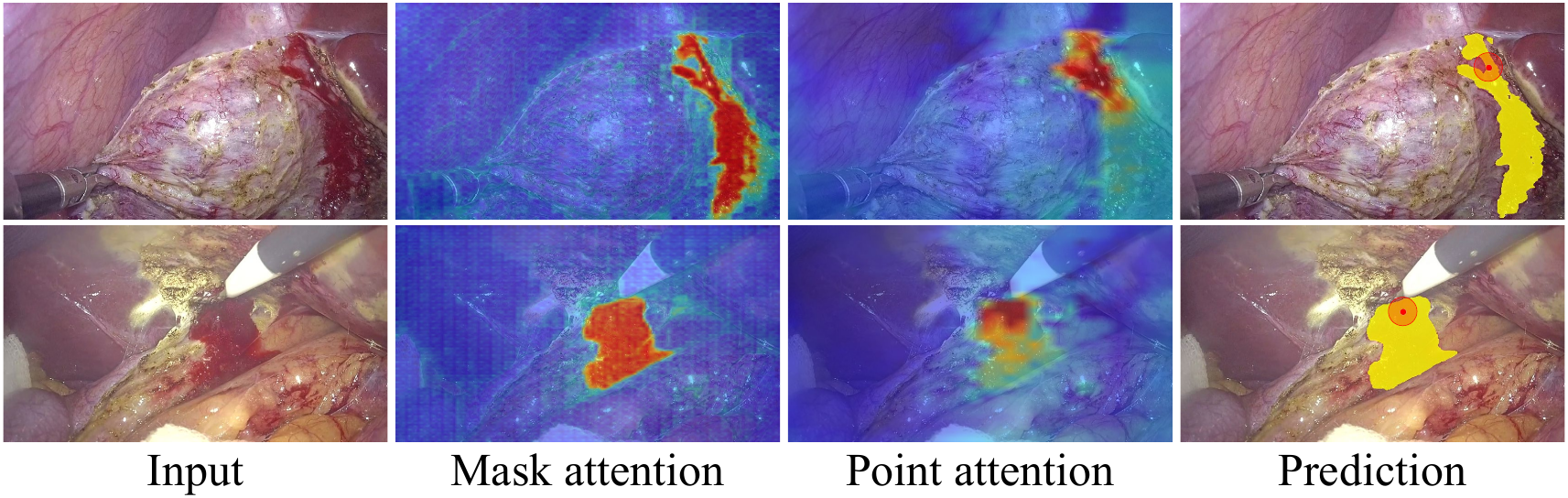}
\vspace{-5pt}
\caption{Attention maps in mask and point branches of~\ourmodel.}
\label{ablat_attn}
\vspace{-5pt}
\end{figure}

\begin{figure}[t!]
\centering
\includegraphics[width=0.99\linewidth]{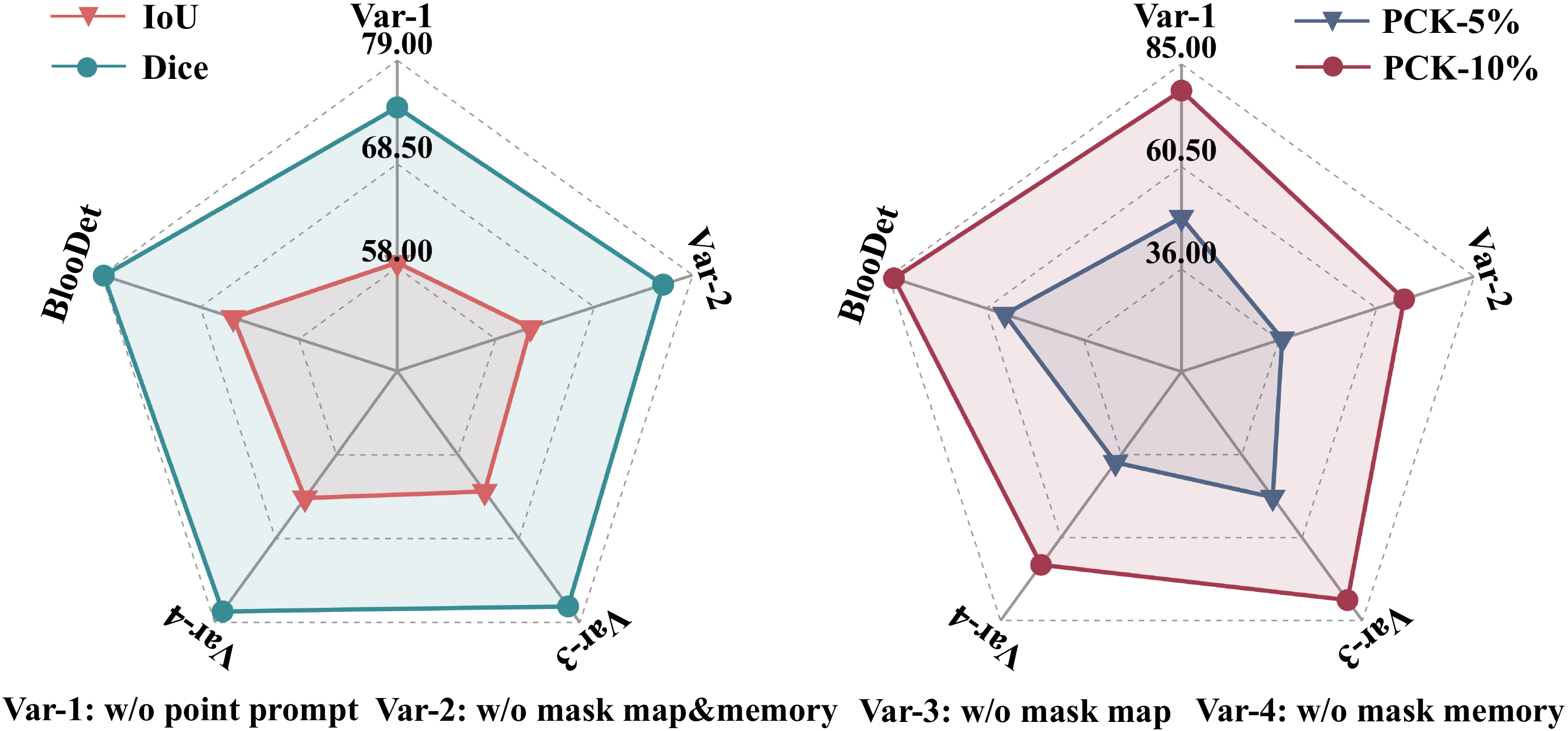}
\vspace{-5pt}
\caption{Effect of mutual guidance between two branches.}
\label{ablat_guide}
\vspace{-10pt}
\end{figure}

\noindent\textbf{Effect of Mutual Guidance.}
To validate the effect of our cross-branch mutual guidance, \figref{ablat_guide} ablates the impact of point prompt from point branch as well as mask map and mask memory from mask branch.
The results indicate that the point prompt contributes to bleeding region detection.
For the point branch, the mask maps from previous frames assist with optical flow operation, while mask memory fosters point memory modeling to improve the accuracy of point localization.
The attention maps visualized in~\figref{ablat_attn}~also verify the effect of mutual guidance.

%% file: sec/6_conclusion.tex
\section{Conclusion}

This work advances the intelligent bleeding region and point detection in laparoscopic surgery.
We contribute an open-source dataset for bleeding detection in real-world surgical videos to facilitate benchmark construction.
Accordingly, we design a dual-task synergistic online framework called~\ourmodel, which assembles mask and point branches in a bidirectional guidance structure, and exploits an edge generator and point memory modeling to enhance the adaptive prompting scheme.
Extensive experimental results demonstrate that our method outperforms existing models for both bleeding tasks.
We believe this study can facilitate research in intelligent surgical assistance, enhancing intraoperative decision-making and clinical outcomes.